\documentclass[10pt, a4paper]{article}

\usepackage[final]{lrec2026} 

\usepackage{times}
\usepackage{boxedminipage}
\usepackage{latexsym}
\usepackage{adjustbox}
\usepackage{listings}
\usepackage{fancyvrb}
\usepackage{booktabs}
\usepackage{csquotes}
\usepackage{multirow}
\usepackage[table,xcdraw,dvipsnames]{xcolor}

\usepackage[T2A,T1]{fontenc}

\usepackage[utf8]{inputenc}
\usepackage{enumitem}

\newcommand{\comment}[1]{}

\newcommand{\enExample}[1]{\noindent \underline{Example:} 
 {\color{blue}\textit{#1}}}

\newcommand{\annotated}[1]{\textbf{#1}}

\newcommand{\slavicExample}[1]{\noindent \underline{Example:} 
 {\color{brown}\textit{#1}}}

\newcommand{\cyrillicExample}[1]{\noindent \underline{Example:} 
    {\color{brown}{\fontencoding{T2A}\fontfamily{cmr}\selectfont\textit{#1}}}}

\newcommand{\translation}[1]{({\color{blue}\textit{#1}})}

\newenvironment{itemizerCompact}{\vspace{-1mm}
  \begin{itemize}
    \setlength{\itemsep}{2pt}
    \setlength{\parskip}{0pt}
    \setlength{\parsep}{0pt}
  }
{ \end{itemize}
  \vspace{-1mm}  }

\title{A Corpus of Persuasion Techniques in Slavic Languages}

\name{{\begin{tabular}{c} 
            Jakub Piskorski$^\star$, 
            Dimitar Iliyanov Dimitrov$^\dagger$, 
            Marina Ernst$^\ddagger$, \\
            Jacek Haneczok$^\diamondsuit$, 
            Michał Marcińczuk$^\spadesuit$, 
            Arkadiusz Modzelewski$^\S$, \\
            Roman Yangarber$ ^\heartsuit$\end{tabular}
    }}

\address{$ ^\star$Institute of Computer Science, Polish Academy of Science, Poland, \\
            $^\dagger$Sofia University "St. Kliment Ohridski", Bulgaria, 
            $^\ddagger$University of Koblenz, Germany, \\
            $^\diamondsuit$Visa Technology Europe, 
            $^\spadesuit$CodeNLP, \\
            $^\S$University of Padua, Italy,
            $ ^\heartsuit$University of Helsinki, Finland 
            \\
            jpiskorski@gmail.com,
            ilijanovd@fmi.uni-sofia.bg,
            marinaernst@uni-koblenz.de, \\
            jacek.haneczok@gmail.com,
            marcinczuk@gmail.com,
            contact@amodzelewski.com,\\
            roman.yangarber@helsinki.fi 
            \\
         }

\abstract{
Persuasion techniques are powerful rhetorical devices used to sway public opinion in a wide range of media.  We present a new corpus of persuasion techniques, focusing on Slavic languages.  The corpus contains documents 
in Bulgarian, Polish, and 
Russian, annotated with persuasion techniques at the coarse-grained text-span level and fine-grained sentence level.  The techniques are drawn from a taxonomy of 25 fine-grained persuasion techniques, grouped under six broad categories of rhetorical persuasion strategies.  The corpus contains approximately 7500 text spans 
from 222 documents that cover topics hotly debated at the national and international levels.  We describe the corpus creation process, provide detailed statistics, and examine correlations between topics and persuasion techniques.  We use classic ML-based and generative AI-based models to provide baselines and benchmark results for the detection and classification of persuasion techniques at the text-span level and sentence level.
 \\ \newline }

\begin{document}

\maketitleabstract

\comment{
******************* TODO *************************

CAMERA-READY 9 pages max

- ??? define TEXT-SPAN

APPENDIX : 10 pages MAX
- full detailed definitions of persuasion technoques 
  - copy-paste from SHARED TASK paper
- models that use LLMs -- provide prompts

}

\section{Introduction}
\label{sec:intro}


Persuasion is central to political debates, affecting policy outcomes, and is heavily used in a wide range of contexts, including by social media influencers, to sway public opinion.  Persuasion techniques function as psychological mechanisms designed to steer the reader's beliefs and actions, such as voting.  Many are based on flawed or unsound reasoning in constructing arguments, while others deliberately trigger emotional responses---such as {\em appeal to patriotism}---to secure agreement when factual support is missing or weak.




In this paper, we present a new corpus of texts annotated with persuasion techniques in Slavic languages.  It contains documents from parliamentary debates in Bulgarian and Polish, and social media in Russian, annotated with persuasion techniques, using a taxonomy of 25 fine-grained persuasion techniques.  The corpus consists of approximately 7,500 text fragments annotated with persuasion techniques, in 222 documents. 
The corpus covers highly debated topics at the international---e.g., the Ukraine-Russia war---and national level---e.g., abortion legislation.

The main impetus behind the creation of the presented corpus is to foster and stimulate research on the detection of persuasion techniques and the classification of texts in Slavic languages.  We aim to cover the domains of parliamentary debates and social media, for which there are few or no resources on persuasion techniques in Slavic languages.  Our contributions can be summarized as follows:

\begin{itemizerCompact}
\item we release a new corpus of persuasion techniques, annotated at the text-span and sentence level in three Slavic languages---Bulgarian, Polish, and Russian---for two text genres: parliamentary debates and social media posts,

\item we provide the characteristics of this corpus, including statistics, topics, and correlations among the labels,

\item we release and evaluate classical ML- and generative-AI-based baseline models for detection and classification of persuasion techniques at the text-span and sentence levels.
    
\end{itemizerCompact}

It is important to emphasize that text-span annotations of persuasion techniques may cover single words, sub-parts of a sentence, entire sentences or even bigger chunks of text that span more than one sentence. The sentence-level annotations are obtained by mapping the text-span annotations to sentences.

The paper is organized as follows. Section~\ref{sec:related} covers related work. Section~\ref{sec:corpus} describes the creation of the corpus\comment{, including: taxonomy, formats, document acquisition, annotation process, and corpus statistics}. Section~\ref{sec:correlation} presents insights into correlations between topics and persuasion techniques. Section~\ref{sec:models} introduces baseline models for persuasion technique detection and classification\comment{, encompassing both classical- and generative AI-based models}.  Section~\ref{sec:conclusions} concludes and outlines future work.

\section{Related Work}
\label{sec:related}



Research on automated detection of five types of logical fallacies was reported by~\citet{Habernal.et.al.2017.EMNLP,Habernal2018b}. \citet{EMNLP19DaSanMartino} presented   a corpus of English news articles labeled using a more fine-grained taxonomy of 18 persuasion techniques at span and sentence level, and reported 
on experiments of automated solutions to detect them. This corpus was used in \textit{NLP4IF-2019 Shared Task on Fine-Grained Propaganda Detection}~\cite{da-san-martino-etal-2019-findings} and \textit{SemEval-2020 Task 11 on Detection of Persuasion Techniques in News Articles}~\cite{da-san-martino-etal-2020-semeval}, focused on the detection of persuasion techniques in text fragments and document-level classification, with an initial taxonomy of 18 techniques, in English news articles. 

\citet{piskorski-etal-2023-multilingual} introduced an extended taxonomy of 23 persuasion techniques, grouped in 6 different categories, and a corpus of 1.7K news articles in 6 languages (including two Slavic languages, Polish and Russian) annotated at the text-span level using this taxonomy. The usefulness and applicability of this taxonomy was demonstrated by \citet{modzelewski-etal-2025-pcot}, who employed it to enhance disinformation detection through the Persuasion-Augmented Chain-of-Thought method. The corpus was used in {SemEval-2023 Task 3 on Detecting Category, Framing, and Persuasion Techniques in Online News in a Multi-lingual Setup}~\cite{piskorski-etal-2023-semeval}. The {CLEF 2024 Task 3 on Persuasion Techniques}~\cite{DBLP:conf/clef/PiskorskiSA0DJP24} followed up on Semeval-2023 Task 3, by including new articles in five languages, namely, Arabic, \textit{Bulgarian}, English, Portuguese, and \textit{Slovene}, (with two additional Slavic languages), and evaluation of persuasion detection and classification at text-span level. 

Recently, work has been reported on detection and classification of persuasion techniques in other text genres, e.g., the shared task \textit{DIPROMATS 2023}: automatic detection and characterization of propaganda techniques in messages from diplomats and authorities of world powers~\cite{moral2023overview} and {DIPROMATS 2024: Detection, characterization and tracking of propaganda in messages from diplomats and authorities of world powers}~\cite{moral2024overview}, which resulted in a release of a dataset of approximately 21K tweets in English and Spanish, posted by authorities of China, Russia, the United States, and the European Union.

Detection and classification of persuasion techniques in parliamentary debates and social media was the subject of the 
\textit{{S}lavic{NLP} 2025 Shared Task}\comment{Detection and Classification of Persuasion Techniques in Parliamentary Debates and Social Media"}~\cite{piskorski-etal-2025-slavicnlp}. In particular, the task focused on the classification of persuasion techniques at the paragraph level, covered 5 Slavic languages (Bulgarian, Croatian, Polish, Russian, and Slovene), and adapted and extended the persuasion technique taxonomy presented in \citet{piskorski-etal-2023-multilingual} to 25 techniques.  Our new dataset, presented in this paper, has been annotated using the same taxonomy. 

\citet{kyslyi-etal-2025-unlp} reports on a shared task on manipulation detection and classification in Ukrainian social media, for which a new dataset containing approximately 9.5K  posts from Ukrainian Telegram in both Russian and Ukrainian were introduced, and labeled with a subset variant of the persuasion taxonomies used in the SemEval tasks mentioned earlier. Moreover, \citet{modzelewski-etal-2024-mipd} proposed a taxonomy comprising 11 manipulation techniques and released a Polish disinformation dataset annotated based on this taxonomy. Finally, \citet{alzahrani2024investigatingpersuasiontechniquesarabic} presented a study on the detection and classification of persuasion techniques in Arabic social media.

Another line of research focused on detecting persuasion techniques in multimodal content. For instance, detection of persuasion techniques in memes was the subject of the shared tasks: \textit{SemEval-2021 Task 6 on Detection of Persuasion Techniques in Texts and Images}~\cite{dimitrov-etal-2021-semeval} and \textit{SemEval-2024 Task 4 on Multilingual Detection of Persuasion Techniques in Memes}~\cite{dimitrov-etal-2024-semeval}. These tasks extended the Semeval-2020 taxonomy to detect persuasion in visual content as well, totaling 22 techniques (20 multimodal and 2 vision-only). Apart from English, the second task covered two Slavic languages: Bulgarian and Macedonian.  

Accurate analysis of persuasion techniques is critical for text-understanding tasks; for example, in information extraction~\cite{piskorski-2013-information-extraction,huttunen-2002-diversity}, persuasion techniques introduce the complex interplay between the factual and the rhetorical dimensions of the content: does the content convey objective reporting of events or attempt to influence the reader's perception of events.  Relying on event detection methods alone is insufficient to assure reliability and trust \cite{atkinson-yangarber:2011:security-springer,yangarber:2006-iiia-verification}.  As persuasive content becomes ever more pervasive online, our ability to accurately detect it and analyze all of its aspects becomes more important.

To the best of our knowledge, the corpus we present is the first for Bulgarian and Polish that covers text-span and sentence-level annotations in the domain of parliamentary debates, and the first for Russian social media, annotated with fine-grained persuasion techniques.  This constitutes part of the corpus used for the shared task described in~\cite{piskorski-etal-2025-slavicnlp}, which focused on paragraph-level detection and classification of persuasion techniques in Slavic languages.






\section{Corpus Creation} 
\label{sec:corpus}

\subsection{Taxonomy} 
\label{sec:taxonomy}

To label the texts in our corpus, we use the 2-tier taxonomy from the SlavicNLP 2025 shared task on persuasion techniques~\cite{piskorski-etal-2023-multilingual}. At the top level are 6 coarse-grained {\em persuasion strategies}: \emph{Attack on Reputation}, \emph{Justification}, \emph{Simplification}, \emph{Distraction}, \emph{Call}, and \emph{Manipulative Wording}---described in detail below.

\comment{
In our analysis, a \textit{persuasion strategy} refers to a high-level category encompassing several persuasive techniques. The dataset includes six distinct persuasion strategies: \textit{Attack on Reputation} [AR], \textit{Justification} [J], \textit{Simplification} [S], \textit{Distraction} [D], \textit{Call} [C], and \textit{Manipulative Wording} [MW].
}


\noindent \textbf{Attack on reputation:} The argument shifts the focus---from addressing the topic, toward targeting the participants---their personality, experience, deeds, etc.---in order to question or undermine their credibility.   The target of the argument can be a group of individuals, organization, \comment{???an object,} or activity.


\noindent \textbf{Justification:} The argument has two parts: a statement and an explanation or appeal, where the latter is used to justify or support the statement.

    
\noindent \textbf{Simplification:} The argument excessively simplifies a problem, usually regarding the cause, the consequence, or the existence of choices.

    
\noindent \textbf{Distraction:} The argument takes focus away from the main topic to distract the reader.


\noindent \textbf{Call:} The text is not an argument, but an encouragement to act or to think in a particular way.


\noindent \textbf{Manipulative wording:} the text is not an argument per se, but uses specific language, which contains words or phrases that are either non-neutral, confusing, exaggerating, loaded, etc., to impact the reader emotionally.


These six strategies, subdivided into 25 fine-grained persuasion techniques (see Figure~\ref{fig:fine-grained-pts}), are defined in detail with examples in Annex~\ref{pt-definitions}.

\begin{figure}
\begin{center}
\begin{boxedminipage}{0.99\linewidth}
\begin{small}
\begin{Verbatim}[commandchars=\\\{\}]
{\bf ATTACK ON REPUTATION} 
- Name Calling or Labelling
- Guilt by Association
- Casting Doubt
- Appeal to Hypocrisy
- Questioning the Reputation

{\bf JUSTIFICATION}
- Flag-Waiving, appeal to patriotism
- Appeal to Authority
- Appeal to Popularity
- Appeal to Fear, Prejudice
- Appeal to Values

{\bf DISTRACTION}
- Strawman
- Whataboutism
- Red Herring
- Appeal to Pity 

{\bf SIMPLIFICATION}
- Causal Oversimplification
- False Dilemma or No Choice
- Consequential Oversimplification
- False Equivalence

{\bf CALL}
- Slogans
- Conversation Killer
- Appeal to Time

{\bf MANIPULATIVE WORDING}
- Loaded Language
- Obfuscation, Intentional Vagueness, 
  Confusion
- Exaggeration or Minimisation
- Repetition
\end{Verbatim}
\end{small}
\end{boxedminipage}
\caption{Two-tier Persuasion Technique taxonomy.}
\label{fig:fine-grained-pts}   
\end{center}
\end{figure}

\subsection{Document Acquisition} 
\label{sec:doc-acquisition}


To create the corpus, we collected documents in three languages---Bulgarian, Polish, and Russian---covering various controversial topics.  For Bulgarian and Polish, we use transcripts of the respective parliamentary sessions.  For Russian, we use social media posts, particularly from the Telegram platform,\comment{excluding official media outlets and instead} focusing on community-based channels. 

Topics covered in Bulgarian parliamentary debates include: foreign policy (with a focus on military aid to Ukraine), acceptance into the Eurozone and Schengen, and national sovereignty concerns.  Domestic political discourse (priorities, integrity, identity) is intertwined with international matters (e.g., ongoing conflicts).

Topics covered in the Polish debates cover the highly-disputed abortion legislation, national security and defense policy, Poland's role within the European Union, legislation against hate speech and discrimination, socio-economic matters---vaccination, forest management, mass layoffs, mental health awareness, etc.

The Russian documents focus predominantly on the Ukraine-Russia war, e.g., Putin-Trump negotiations, Russia's opposition with the West, disinformation, demographic challenges (such as migration, integration), criticism of state policies in this context, and civilian resistance in conflict zones.


All documents are multiply labeled with broad public discourse {\em topics}: \textit{Ukraine-Russia war} (URW), \textit{Defense}, \textit{Israel-Hamas conflict}, \textit{Migration}, \textit{Demographics}, \textit{Abortion}, \textit{EU}, \textit{Schengen}, \textit{External affairs}, \textit{Climate change}, \textit{Hate speech}, \textit{History}, \textit{Other}.

\subsection{Annotation Process}
\label{sec:annotation-process}

The annotation process included five steps:

\begin{enumerate}[nosep, leftmargin=*]

\item For each language, a team of at least two annotators, one curator, and one language coordinator was set up, where all team members were native speakers and had prior experience in linguistic annotations, including annotation in disinformation and manipulative content,

\item We provided all teams with detailed annotation guidelines~\cite{JRC132862}, and organized live sessions with annotation trainings,

\item Each document was annotated by two annotators; ad-hoc meetings were organized to discuss difficult and unclear cases,

\item The curator for each language verified the adherence of the annotations to the guidelines, and corrected them as needed,

\item Regular meetings were held and an information exchange platform was set up to safeguard and maintain consistency in annotations across languages, given the complexity of the task at hand~\cite{stefanovitch-piskorski-2023-holistic}.

\end{enumerate}

We adapted INCEpTION~\cite{tubiblio106270}---a web-based framework for collaborative annotation.\comment{ A screenshot of the interface of INCEpTION is provided in Figure~\ref{fig:inception}.

\begin{figure}
    \centering
    \includegraphics[width=0.99\linewidth]{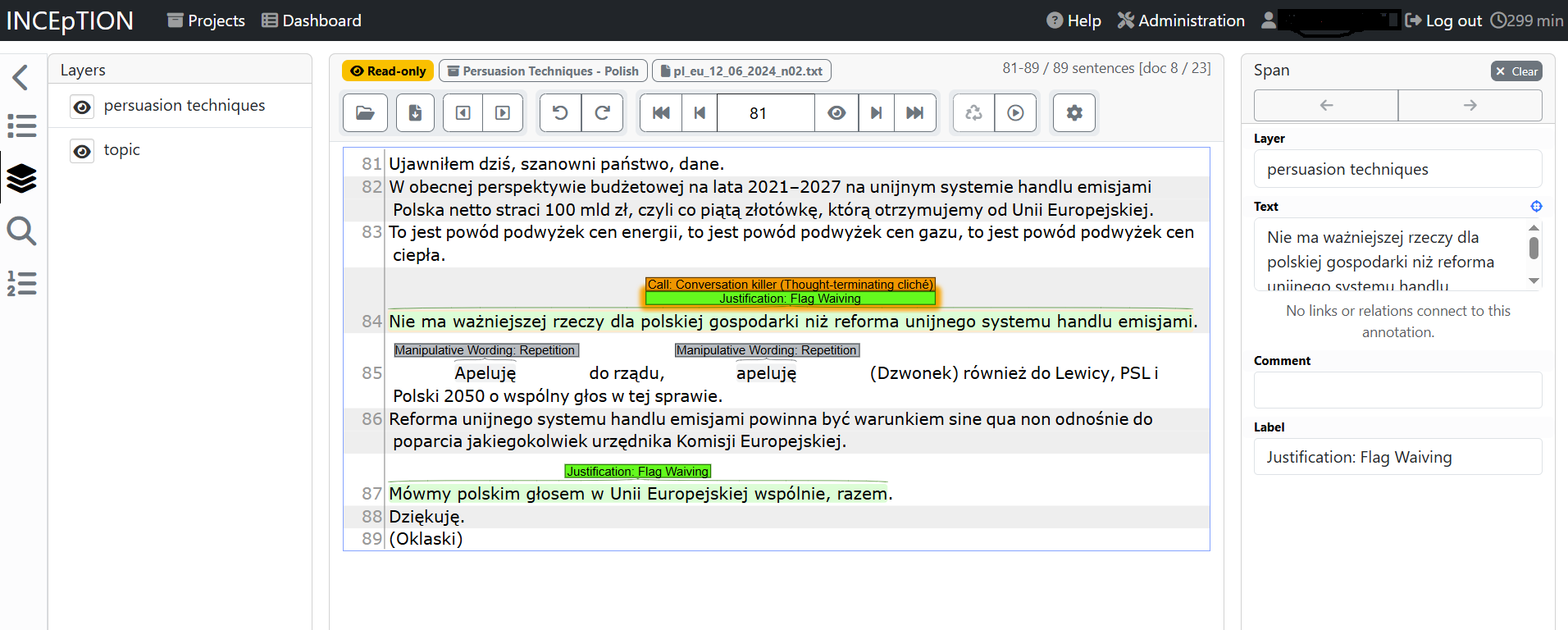}
    \caption{Screenshot of the INCEpTION platform.}
    \label{fig:inception}
\end{figure}
}


In the remainder of this section, we outline the types of persuasion techniques that proved most challenging for the annotators to detect reliably. Their subtlety lies not in overt manipulation but in how they blend into everyday reasoning, requiring the reader to track context, nuance, and argument structure to recognize them. 

{\em Whataboutism} and {\em red herrings} are particularly tricky in complex debates, where topic-switching feels relevant.  They rely on the reader losing track of the original issue as the discussion expands.  Spotting these techniques requires considering the broad context and noticing when the focus has shifted without justification.  {\em False equivalence} can be equally deceptive, as it presents opposing ideas in a ``balanced'' format that appears fair, requiring a deep understanding of the underlying issues to see that the comparison is flawed.  {\em Strawman} arguments distort the opposing position slightly, often in long or nuanced exchanges; detecting them requires remembering what was said earlier and recognizing subtle misrepresentations.  Causal or consequential {\em oversimplification} appeals to our desire for clear, linear explanations, though real-world problems often involve multiple causes, feedback loops, and uncertainties. Identifying this type of fallacy requires a cognitive effort to weigh the missing variables and alternative outcomes.  Finally, {\em obfuscation}, intentional vagueness, and confusion hide flawed reasoning behind abstract or technical language; the reader feels something meaningful has been said but can’t specify what.  Detecting this tactic requires noticing the absence of clarity rather than the presence of persuasion, i.e., it is a task that demands sustained attention and critical reflection across the broader context.

\subsection{Annotation Format}
\label{sec:annotation-format}

For each language in the dataset, there is one annotation file, where each line contains annotations at \textbf{sentence level} and has the following format:

\begin{verbatim}
<FILE> <START> <END> <PT-LABELS>
\end{verbatim}

\noindent where \verb+<FILE>+, is the name of the file, \verb+<START>+
and \verb+<END>+ are the start/end character position of the sentence, and \verb+<PT-LABELS>+ is a list of persuasion technique labels. If the given sentence does not contain any persuasion techniques, then the list is empty. All elements in each line are separated by tabs. 

For each language, the dataset contains a corresponding file with \textbf{single text-span annotations}, where each line has the format:

\begin{verbatim}
<FILE> <START> <END> <PT> <NORM>
\end{verbatim}

\noindent where \verb+<PT>+ contains a unique persuasion technique label and 
\verb+<NORM>+ contains whitespace-normalized text which corresponds to the text fragment starting/ending at \verb+<START>+/\verb+<END>+. 

All documents in the corpus are provided in plain text format encoded in UTF-8.



\subsection{Statistics}
\label{sec:statistics}

\begin{figure*}[ht]
    \centering
    \includegraphics[trim=11mm 5mm 12mm 25mm, clip, width=0.99\linewidth]{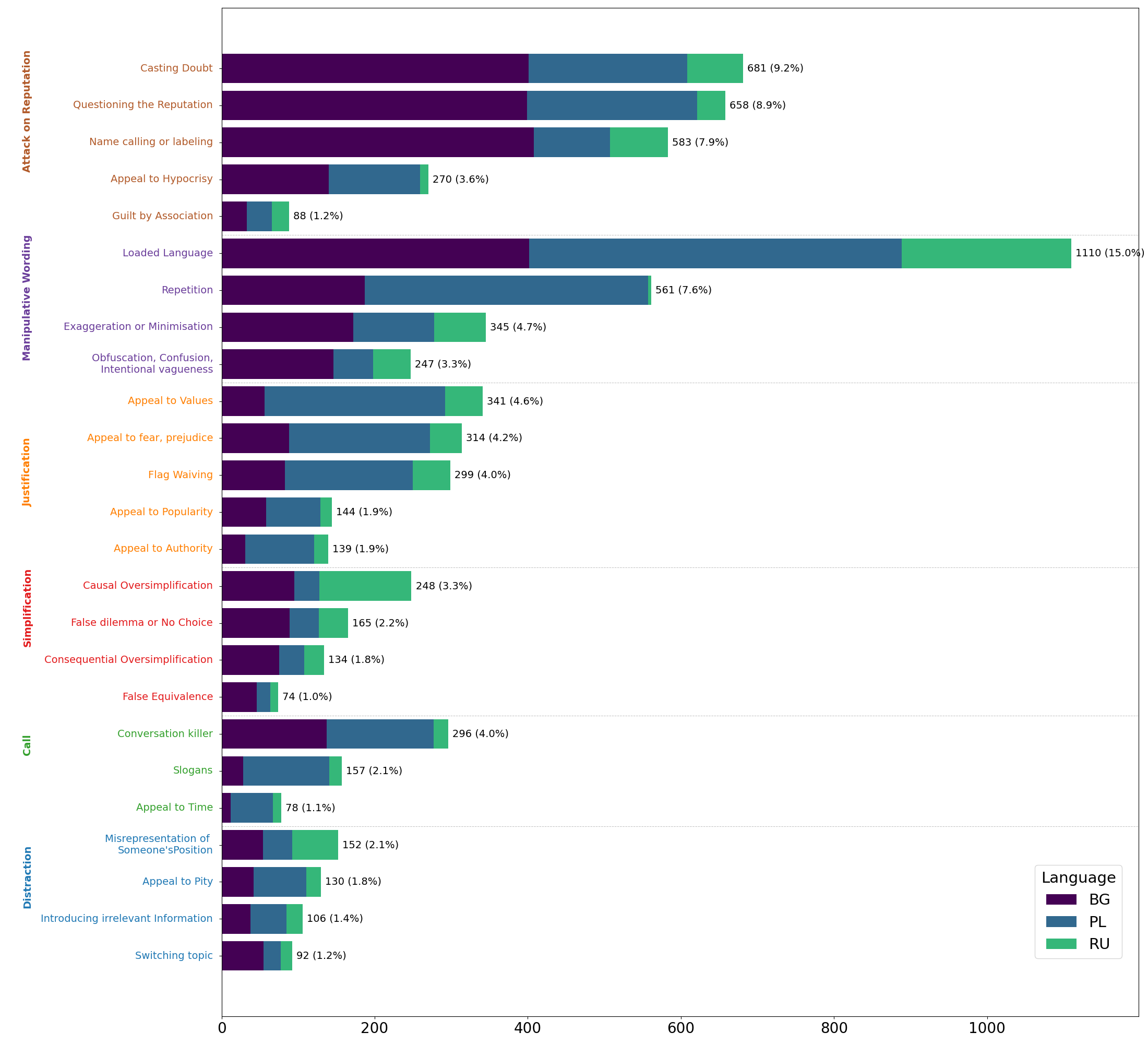}
    \caption{Distribution of persuasion technique annotations by language.}
    \label{fig:pt-distro}
\end{figure*}

The corpus consists of 222 documents in three languages---Bulgarian, Polish, and Russian---with a total of 7412 instances of persuasion techniques. 
The overall statistics of the corpus are provided in Table~\ref{tab:data-stats}.
The proportion of sentences labeled with at least one persuasion technique for Bulgarian, Polish, and Russian is 50.0\%, 46.3\%, and 58.7\%, respectively. The dataset covers 25 persuasion techniques, with a highly unbalanced distribution. Figure~\ref{fig:pt-distro} provides the distribution of the persuasion techniques and comparison across languages. \textit{False Equivalence} is the least frequent class (74 instances), while \textit{Loaded language} is the most frequent (1110 instances).


\begin{table}[t]
\centering
\begin{scriptsize}
\resizebox{.99\columnwidth}{!}{%
\begin{tabular}{p{11em}rrrrr}
\hline
 & \textbf{BG} & \textbf{PL} & \textbf{RU} \\
\toprule
Documents & 79 & 53 & 90 \\
Sentences & 4984 & 4515 & 2074 \\
Sentences with PTs & 2491 & 2092 & 1218 \\
Text spans annotated & 3275  & 3048 & 1089 \\
AVG words/document & 1146 & 1215 & 359 \\
\bottomrule
\end{tabular}
}
\end{scriptsize}
\caption{Dataset statistics across languages.}
\label{tab:data-stats}
\end{table}

\begin{figure*}[ht]
    \centering
    \includegraphics[trim=20mm 8mm 6mm 5mm, clip, width=0.99\linewidth]{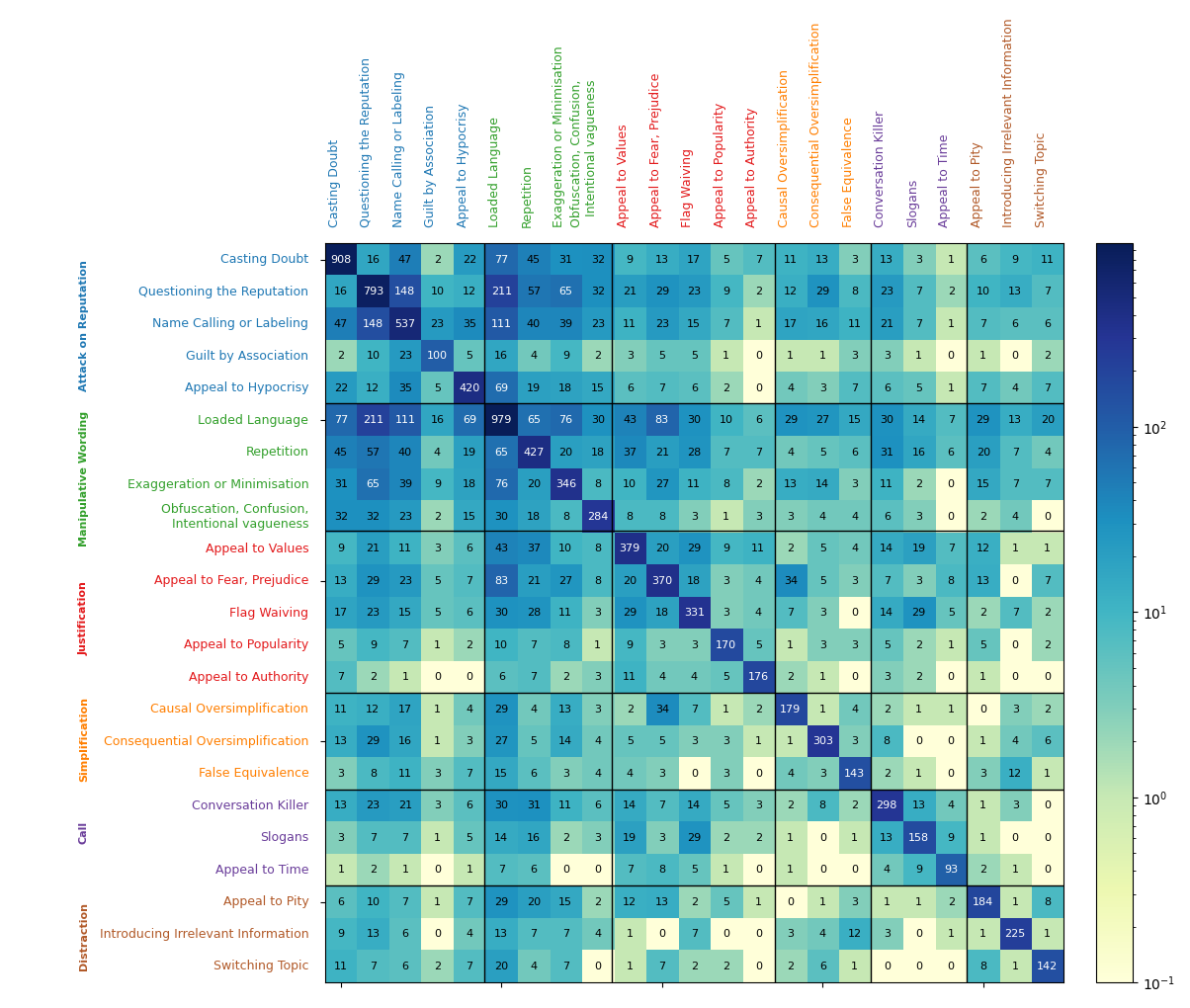}
    \caption{Co-occurrence of persuasion technique annotations within sentences.}
    \label{fig:pt-matrix}
\end{figure*}

The heatmap in Figure~\ref{fig:pt-matrix} shows the co-occurrence of persuasion-technique annotations within sentences across all languages.
Each cell shows the number of sentences that contain both persuasion techniques corresponding to the row and the column. The diagonal cells indicate the total number of sentences annotated with a specific persuasion technique. Darker colors indicate higher counts, highlighting techniques that frequently recur within the same sentence. The axes list all persuasion techniques, grouped and color-coded by their broader categories (e.g., \emph{Attack on Reputation}, \emph{Manipulative Wording}, \emph{Justification}, \emph{Simplification}, \emph{Call}, \emph{Distraction}).

\comment{
  \begin{table}[t]
\centering
\resizebox{.99\columnwidth}{!}{%
\begin{tabular}{p{9.5em}|rr|rr|rr|rr}
\hline
{\bf Topic} & \textbf{BG} && \textbf{PL} && \textbf{RU}  &    \\
\toprule
Abortion              &  - &        & 14 &  26\% &  1 &    1\% \\
Climate change        &  2 &  2.5\% &  8 &  15\% &  - &        \\
Defense               & 14 &   17\% & 22 &  41\% &  - &        \\
Demographics          &  1 &  1.2\% &  5 &   9\% &  9 &   10\% \\
Hate speech           &  - &        &  8 &  15\% &  - &        \\
EU                    & 28 &   35\% & 15 &  28\% &  - &        \\
External affairs      &  2 &  2.5\% & 13 &  25\% & 21 &   23\% \\
History               &  - &        &  2 &   4\% &  - &        \\
Israel-Hamas conflict &  2 &  2.5\% &  - &       &  - &        \\
Migration             &  7 &  8.5\% &  8 &  15\% & 13 & 14.5\% \\
Schengen              &  7 &  8.5\% &  - &       &  - &        \\
Ukraine-Russia war    & 42 &   52\% &  5 &   9\% & 52 &   58\% \\
Other                 &  4 &    5\% &  1 &   2\% &  4 &  4.5\% \\

\bottomrule
\end{tabular}
}
\caption{Topic distribution across languages.}
\label{tab:topic-stats}
\end{table}
}

\begin{table}[t]
\centering
\resizebox{.99\columnwidth}{!}{%
\begin{tabular}{p{9.5em}|rr|rr|rr|rr}
\hline
{\bf Topic} & \textbf{BG} & \%\% & \textbf{PL} & \%\% & \textbf{RU}  & \%\%   \\
\toprule
Abortion              &  - &      & 14 &  26 &  1 &    1 \\
Climate change        &  2 &  2.5 &  8 &  15 &  - &      \\
Defense               & 14 &   17 & 22 &  41 &  - &      \\
Demographics          &  1 &  1.2 &  5 &   9 &  9 &   10 \\
Hate speech           &  - &      &  8 &  15 &  - &      \\
EU                    & 28 &   35 & 15 &  28 &  - &      \\
External affairs      &  2 &  2.5 & 13 &  25 & 21 &   23 \\
History               &  - &      &  2 &   4 &  - &      \\
Israel-Hamas conflict &  2 &  2.5 &  - &     &  - &      \\
Migration             &  7 &  8.5 &  8 &  15 & 13 & 14.5 \\
Schengen              &  7 &  8.5 &  - &     &  - &      \\
Ukraine-Russia war    & 42 &   52 &  5 &   9 & 52 &   58 \\
Other                 &  4 &    5 &  1 &   2 &  4 &  4.5 \\

\bottomrule
\end{tabular}
}
\caption{Topic distribution across languages.}
\label{tab:topic-stats}
\end{table}

Table~\ref{tab:topic-stats} summarises the distribution of documents by topic across the Bulgarian (BG), Polish (PL), and Russian (RU) datasets. The figures reveal significant cross-linguistic variation. All three languages engage with the \verb+URW+ topic, which accounts for 44\% of documents across languages, though it is particularly prevalent in Russian (58\%) and Bulgarian (52\%) texts. \verb+External affairs+ emerges as another cross-linguistic theme, particularly relevant in Russian (23\%) and Polish (25\%) materials. In contrast, while \verb+EU+ dominates Bulgarian texts (35\%), \verb+Abortion+ debates appear almost exclusively in Polish discourse (26\%) 


\section{Topic vs. Persuasion Technique Analysis}
\label{sec:correlation}

\begin{figure*}[ht]
    \centering
    \includegraphics[trim=2.5mm 15mm 3mm 3mm, clip, width=0.99\linewidth]{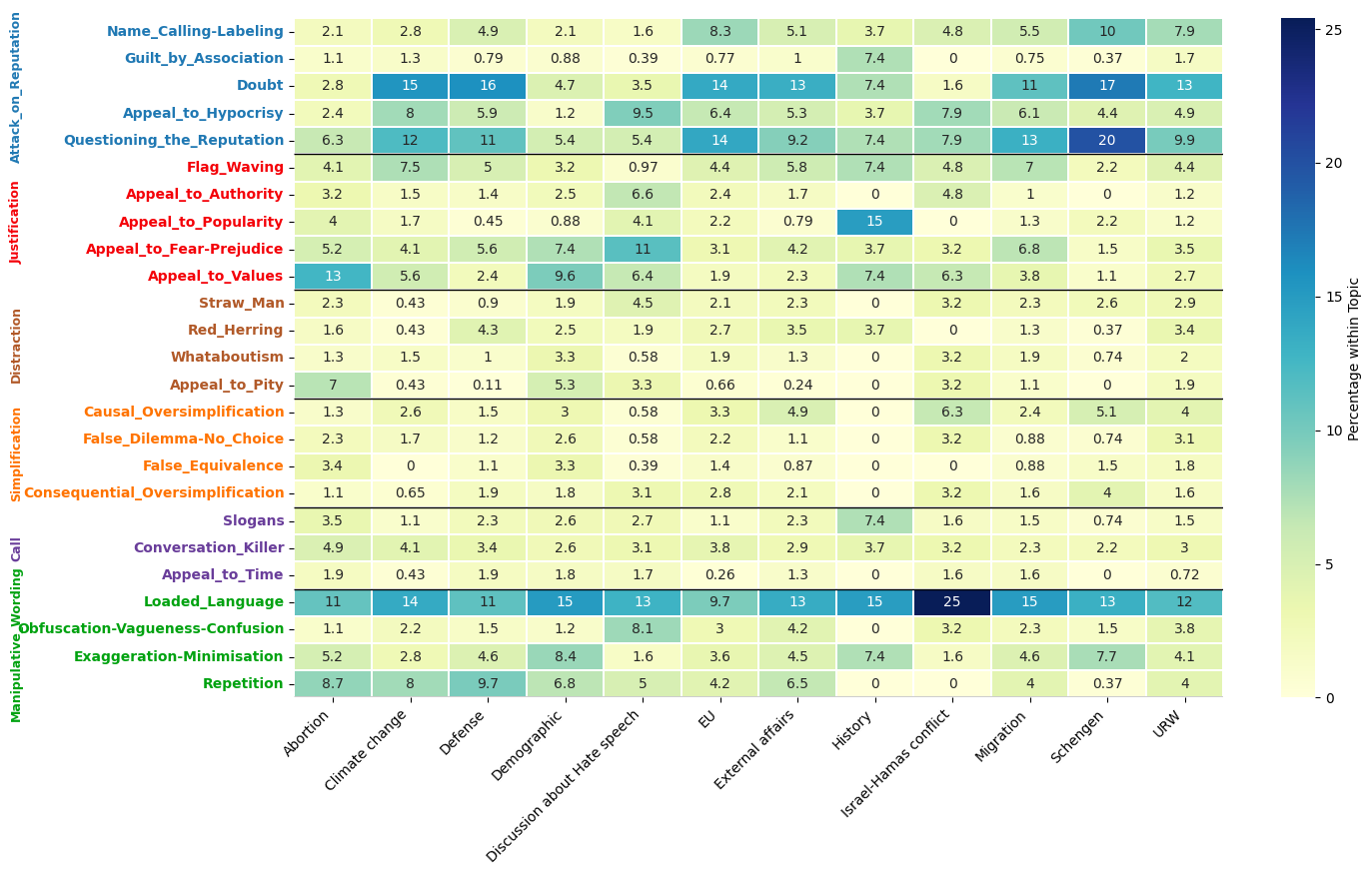}
    \caption{Correlation between persuasion technique and topics across languages.}
    \label{fig:topics-pts}
\end{figure*}

To better understand how different persuasion techniques and strategies are employed in public discourse, we analyze their co-occurrence with the topics under discussion. 
Figure ~\ref{fig:topics-pts} presents a heatmap of persuasion-topic co-occurrence: the rows represent persuasion techniques, grouped into strategies, and the columns represent topics. Color intensity reflects the prevalence of each strategy within a topic, while the numerical values provide precise percentages.

Across all topics, \emph{Manipulative Wording} and \emph{Attack on Reputation} are the most prominent persuasion strategies, with \textit{Loaded Language} being the most prevalent technique. \textit{Doubt} and \textit{Questioning the Reputation} frequently occur in politically sensitive discussions, particularly those related to \verb+EU+, \verb+Defense+, \verb+Schengen+, \verb+Migration+.

Several techniques exhibit topic-specific spikes. For example, \textit{Appeals to values} occurs more frequently in documents related to \verb+Abortion+ and \verb+Demographic+, whereas \textit{Appeal to Fear} becomes particularly important in \verb+Hate speech+. 
In contrast, \textit{Doubt} and \textit{Questioning the Reputation} appear frequently in documents about \verb+URW+, the most common topic in the corpus. It is important to note that \textit{Loaded language} is a dominant technique in the Russian data, while Bulgarian contributes most with \textit{Doubt}. Notably, the topics \verb+Schengen+ and \verb+Migration+ evoke similar persuasion techniques.

\section{Baseline Models}
\label{sec:models}

In this Section, we present baseline models for the task of detection and classification of persuasion techniques.
Since this paper focuses on the corpus, we provide these models and results as benchmarks (rather than as state-of-the-art)---to highlight the {\em complexity} of the classification task.  

\subsection{Text-span level}

To demonstrate the complexity of the Persuasion Technique Classification problem, we build a baseline model that, given a text fragment known to contain one or more persuasion techniques and no additional context, determines the persuasion technique.  For this purpose, we trained a SVM\footnote{We used the Liblinear library at: \href{https://www.csie.ntu.edu.tw/~cjlin/liblinear/}{\url{www.csie.ntu.edu.tw/~cjlin/liblinear}}} on text spans annotated with persuasion techniques, which uses 3-5 character n-grams as features, with vector normalization, and ran 5-fold cross-validation on the entire data for each language. The classification results (i.e., $F_1$ scores) for individual techniques are provided in Table~\ref{tab:text-span-classification}. The macro (micro) $F_1$ for the persuasion technique classification task is .19 (.34), .18 (.35), and .04 (.11) for Bulgarian, Polish, and Russian, respectively. For the individual techniques, we can observe that lexical features have some discriminative power for the classification of \emph{Loaded Language}, \emph{Name Calling and Labeling}, \emph{Repetition}, \emph{Conversational Killer}, \emph{Flag Waiving}, \emph{Appeal to Values}, where these classes appear to have a higher number of instances in the corpus. On the other hand, \emph{Distractions} and \emph{Simplifications} are less frequent, and intuitively harder to detect. In general, results for Russian are worse than those for Bulgarian and Polish, likely due to the smaller amount of available data for Russian.

\begin{table}[t]
  \centering
     \resizebox{1.02\columnwidth}{!}{%
    \begin{tabular}{@{} lrrrrrr}
      \toprule
      & \multicolumn{2}{c}{\sc BG} & \multicolumn{2}{c}{\sc PL} 
      & \multicolumn{2}{c}{\sc RU} \\
      \midrule
Technique & $F_1$ & sup & $F_1$ & sup & $F_1$ & sup \\
      \midrule   
Questioning the Reputation & .26 & 393 & .19 & 222 & .00 & 37 \\
Name-Calling, Labeling & {\sethlcolor{LimeGreen}\hl{.56}} & 343 & {\sethlcolor{Yellow}\hl{.37}} & 90 & .09 & 76\\
Guilt by Association & .00 & 31 & .00 & 32 & .00 & 22\\
Doubt & {\sethlcolor{GreenYellow}\hl{.40}} & 396 & .19 & 207 & .05 & 73 \\
Appeal to Hypocrisy & .22 & 139 & .09 & 118 & .00 & 11\\
\midrule
Appeal to Popularity & .22 & 57 & .09 & 71 & .00 & 15 \\
Appeal to Values & .03 & 56 & {\sethlcolor{GreenYellow}\hl{.41}} & 236 & .15 & 49\\
Appeal to Authority & .11 & 31 & .28 & 89 & .00 & 18 \\
Appeal to Fear/Prejudice & .10 & 86 & .15 & 184 & .00 & 42 \\
Flag Waving & .24 & 83 & {\sethlcolor{Yellow}\hl{.39}} & 167 & .04 & 49\\
\midrule					  
Causal Oversimplification & .05 & 95 & .00 & 33 & .07 & 120 \\
False Dilemma-No Choice & .10 & 89 & .04 & 38 & .00 & 38 \\
Consequential Oversimplification & .02 & 75 & .00 & 33 & .00 & 26 \\
False Equivalence & .07 & 46 & .00 & 18 & .00 & 10\\
\midrule
Straw Man & .06 & 54 & .00 & 38 & .00 & 60 \\
Whataboutism & .03 & 55 & .21 & 22 & .00 & 15 \\
Red Herring & .05 & 35 & .07 & 47 & .00 & 21 \\
Appeal to Pity & .14 & 42 & .11 & 69 & .00 & 19\\
\midrule
Appeal to Time & .00 & 12 & .14 & 55 & .15 & 11\\
Slogans & .12 & 26 & .12 & 112 & .00  & 16 \\
Conversation Killer & {\sethlcolor{GreenYellow}\hl{.46}} & 129 & {\sethlcolor{Yellow}\hl{.31}} & 136 & .00 & 19 \\
\midrule
Loaded Language & {\sethlcolor{LimeGreen}\hl{.60}} & 363 & {\sethlcolor{LimeGreen}\hl{.68}} & 461 & {\sethlcolor{Yellow}\hl{.32}} & 222 \\
Obfuscation-Vagueness-Confusion & .14 & 145 & .12 & 52 & .00 & 49\\
Exaggeration-Minimization & .14 & 164 & .06 & 106 & .00 & 67\\
Repetition & {\sethlcolor{LimeGreen}\hl{.59}} & 123 & {\sethlcolor{LimeGreen}\hl{.59}} & 298 & .00 & 4\\
      \bottomrule
ALL (micro) & .34 & 3068 & .35 & 2934 & .11 & 1089 \\
ALL (macro) & .19 & 3068 & .18 & 2934 & .04 & 1089 \\
     \bottomrule 
    \end{tabular}
    }
    \caption{Text-span classification---$F_1$ and support---of character n-gram based SVM. Scores above .5, in range .4--.5, and in range .3--.4 highlighted in green, lime green, and yellow, respectively.}
  \label{tab:text-span-classification}
\end{table}


\subsection{Sentence level}
\subsubsection{Coarse-Grained Persuasion Strategy Classification with LLMs}

\begin{table*}
    \centering
    \resizebox{.90\textwidth}{!}{%
    \begin{tabular}{lcccccc}
      \toprule
      & \multicolumn{2}{c}{\sc BG} & \multicolumn{2}{c}{\sc PL} 
      & \multicolumn{2}{c}{\sc RU} \\
      \midrule
Model & $F_1$ Macro & $F_1$ Micro & $F_1$ Macro & $F_1$ Micro & $F_1$ Macro & $F_1$ Micro \\
      \midrule 
gemini-2.0-flash & .28 & .35 & .27 & .34 & .20 & .26 \\
google/gemma-3-27b-it & .27 & \sethlcolor{LimeGreen}\hl{.40} & \sethlcolor{Yellow}\hl{.29} & \sethlcolor{GreenYellow}\hl{.36} & .20 & \sethlcolor{Yellow}\hl{.28} \\
gpt-4.1-mini & .26 & .33 & .27 & .34 & .21 & .25 \\
gpt-4o-mini& \sethlcolor{Yellow}\hl{.29} & .35 & .28 & .32 & \sethlcolor{Yellow}\hl{.23} & .27 \\
meta-llama/Llama-3.3-70B-Instruct & .25 & .33 & .24 & .32 & .22 & \sethlcolor{Yellow}\hl{.28} \\
\bottomrule
\end{tabular}
}
    \caption{Results for multi-label classification of high-level persuasion strategies. Results presented for six different LLMs.  Experiments done at the sentence level with zero-shot prompting. Scores above .5, in range .4--.5, and in range .3--.4 highlighted in green, lime green, and yellow, respectively.}
    \label{tab:multilabel_on_high_level_groups}
\end{table*}

\begin{table*}
    \centering
   \resizebox{.99\textwidth}{!}{%
\begin{tabular}{lrcccccc}
\toprule
 & Model & Attack on Reputation & Call & Distraction & Justification & Manipulative Wording & Simplification \\
\midrule
\multirow[t]{5}{*}{BG} & gemini-2.0-flash & .45 & .18 & .18 & .19 & .47 & .32 \\
 & google/gemma-3-27b-it & \sethlcolor{LimeGreen}\hl{.56} & .16 & .22 & \sethlcolor{Yellow}\hl{.22} & .46 & .16 \\
 & gpt-4.1-mini & .44 & .20 & .26 & .15 & .47 & .36 \\
 & gpt-4o-mini & .45 & \sethlcolor{Yellow}\hl{.23} & \sethlcolor{Yellow}\hl{.28} & .10 & .47 & \sethlcolor{GreenYellow}\hl{.39} \\
 & meta-llama/Llama-3.3-70B-Instruct & .42 & .12 & .08 & .14 & \sethlcolor{LimeGreen}\hl{.48} & .18 \\
 \midrule
\multirow[t]{5}{*}{PL} & gemini-2.0-flash & .40 & .22 & .14 & \sethlcolor{GreenYellow}\hl{.38} & \sethlcolor{LimeGreen}\hl{.50} & .14 \\
 & google/gemma-3-27b-it & \sethlcolor{LimeGreen}\hl{.44} & .10 & .10 & .32 & .47 & \sethlcolor{Yellow}\hl{.28} \\
 & gpt-4.1-mini & .43 & \sethlcolor{Yellow}\hl{.27} & .15 & .28 & .46 & .18 \\
 & gpt-4o-mini & .41 & .20 & .18 & .22 & .48 & .17 \\
 & meta-llama/Llama-3.3-70B-Instruct & \sethlcolor{LimeGreen}\hl{.44} & .15 & \sethlcolor{Yellow}\hl{.22} & .15 & .46 & .22 \\
\midrule
\multirow[t]{5}{*}{RU} & gemini-2.0-flash & .26 & .12 & .08 & \sethlcolor{Yellow}\hl{.24} & .40 & .28 \\
 & google/gemma-3-27b-it & .37 & .12 & .12 & .17 & .33 & .18 \\
 & gpt-4.1-mini & \sethlcolor{LimeGreen}\hl{.46} & \sethlcolor{Yellow}\hl{.18} & .10 & .16 & \sethlcolor{LimeGreen}\hl{.41} & \sethlcolor{Yellow}\hl{.29} \\
 & gpt-4o-mini & .42 & .11 & \sethlcolor{Yellow}\hl{.14} & .00 & .38 & .27 \\
 & meta-llama/Llama-3.3-70B-Instruct & .41 & .12 & .10 & .08 & .40 & .28 \\
\bottomrule
\end{tabular}
}
    \caption{Binary classification results for each of six high-level persuasion strategies. Experiments done at the sentence level with zero-shot prompting. Scores above .5, in range .4--.5, and in range .3--.4 highlighted in green, lime green, and yellow, respectively.}
    \label{tab:binary_on_high_level_groups}
\end{table*}

\newcommand{\myparagraph}[1]{\vspace{2mm}\noindent\textbf{#1}:}
\myparagraph{Experimental Setup}
To evaluate the capability of large language models (LLMs) to detect persuasion at the sentence level, we assess LLM performance on two classification tasks:
\begin{itemize}[nosep, leftmargin=*]
    \item \textbf{Binary Classification} -- Each persuasion strategy is treated as an independent binary classification problem. Models are evaluated on their ability to detect the presence or absence of a specific persuasion strategy.
    \item \textbf{Multilabel Classification} -- LLMs are assessed on their ability to identify multiple persuasion strategies that may co-occur within a single sentence, formulated as a multilabel classification.
\end{itemize}

We evaluate five state-of-the-art LLMs accessed via their respective APIs\footnote{For open-source LLMs we used APIs provided by DeepInfra: \href{https://deepinfra.com/models}{deepinfra.com/models}}: \textit{GPT-4o Mini}, \textit{GPT-4.1 Mini}, \textit{Gemini 2.0 Flash}, \textit{Gemma 3 27B IT}, and \textit{Llama 3.3 70B}. To ensure that the results are as deterministic as possible, all models were prompted with the temperature parameter set to zero.  Evaluations were conducted in a zero-shot setting for all languages.  In total, the experiments involved approximately 55,000 API calls.  The complete set of prompts and the accompanying codebase are available in our public repository to support transparency and reproducibility.\footnote{Link to repository:
\href{https://github.com/SlavicNLP/SlavicPersuasionTechniques/}{github.com/SlavicNLP/\\SlavicPersuasionTechniques/}}

\myparagraph{Results}
Overall performance is moderate and comparable across languages, reflecting the difficulty of zero-shot persuasion detection.  In the multilabel setup (Table \ref{tab:multilabel_on_high_level_groups}), micro $F_1$  ranges from .26 to .40, with Gemma 3 27B achieving the best results, particularly for Bulgarian and Polish.

For the binary classification task (Table \ref{tab:binary_on_high_level_groups}), variation across persuasion strategies is substantial.  The models perform best on \textit{Attack on Reputation} and \textit{Manipulative Wording}, often exceeding the $F_1$ score of .45, while strategies such as \textit{Call} and \textit{Distraction} remain challenging.  Performance is generally consistent across languages, though Russian shows slightly lower scores.

\subsubsection{Fine-Grained Persuasion Technique Classification with Custom SLMs}
\myparagraph{Experimental Setup}
We fine-tuned a transformer-based model for each language---\emph{PKOBP/polish-roberta-8k}\footnote{\href{https://huggingface.co/PKOBP/polish-roberta-8k}{huggingface.co/PKOBP/polish-roberta-8k}} for Polish, \emph{ai-forever/ruRoberta-large}\footnote{\href{https://huggingface.co/ai-forever/ruRoberta-large}{huggingface.co/ai-forever/ruRoberta-large}} \cite{zmitrovich-etal-2024-family} for Russian, and \emph{FacebookAI/xlm-roberta-large}\footnote{\href{https://huggingface.co/FacebookAI/xlm-roberta-large}{huggingface.co/FacebookAI/xlm-roberta-large}} \cite{DBLP:journals/corr/abs-1911-02116} for Bulgarian. Training was conducted for 60 epochs with a learning rate of 5e-6, batch size of 16, maximum sequence length of 512 tokens, and weight decay of 0.01. For each language-specific model, the checkpoint yielding the highest micro $F_1$ score on the validation subset was selected, and final results were reported on the corresponding held-out test set. Table \ref{tab:dataset_stats_sentence_level} presents the sentence-level statistics, showing the number of sentences with each type of PT per language and subset (train, validation, and test).

\begin{table}[t]
  \centering
     \resizebox{.99\columnwidth}{!}{%
                \begin{tabular}{lrrr|rrr|rrr}
                        \toprule
                            & \multicolumn{3}{c|}{\sc BG} & \multicolumn{3}{c|}{\sc PL} & \multicolumn{3}{c}{\sc RU} \\
                        \midrule    
                        Technique & train & valid & test & train & valid & test & train & valid & test \\
                        \midrule
                        Questioning the Reputation & 315 & 139 & 41 & 181 & 45 & 33 & 22 & 13 & 4 \\
                        Name-Calling, Labeling & 233 & 97 & 38 & 66 & 12 & 11 & 44 & 13 & 22 \\
                        Guilt by Association & 24 & 8 & 2 & 28 & 9 & 4 & 13 & 8 & 4 \\
                        Doubt & 316 & 152 & 62 & 221 & 43 & 26 & 47 & 19 & 20 \\
                        Appeal to Hypocrisy & 129 & 73 & 26 & 135 & 25 & 15 & 8 & 1 & 7 \\
                        \midrule
                        Appeal to Popularity & 39 & 22 & 3 & 57 & 21 & 8 & 13 & 1 & 5 \\
                        Appeal to Values & 41 & 14 & 8 & 176 & 60 & 19 & 32 & 11 & 18 \\
                        Appeal to Authority & 35 & 13 & 6 & 71 & 16 & 12 & 12 & 5 & 4 \\
                        Appeal to Fear/Prejudice & 70 & 33 & 13 & 144 & 29 & 32 & 27 & 10 & 12 \\
                        Flag Waving & 67 & 21 & 10 & 124 & 34 & 17 & 32 & 9 & 16 \\
                        \midrule
                        Causal Oversimplification & 80 & 19 & 10 & 31 & 8 & 2 & 90 & 27 & 36 \\
                        False Dilemma-No Choice & 64 & 25 & 12 & 35 & 8 & 3 & 23 & 11 & 17 \\
                        Consequential Oversimplification & 61 & 22 & 15 & 34 & 6 & 6 & 21 & 10 & 3 \\
                        False Equivalence & 49 & 22 & 8 & 34 & 9 & 4 & 10 & 4 & 3 \\
                        \midrule
                        Straw Man & 42 & 19 & 10 & 50 & 12 & 6 & 39 & 22 & 10 \\
                        Whataboutism & 57 & 28 & 8 & 16 & 7 & 3 & 11 & 3 & 5 \\
                        Red Herring & 85 & 30 & 15 & 43 & 12 & 7 & 16 & 7 & 10 \\
                        Appeal to Pity & 31 & 5 & 8 & 86 & 11 & 8 & 13 & 8 & 13 \\
                        \midrule
                        Appeal to Time & 10 & 2 & 1 & 42 & 15 & 8 & 9 & 3 & 3 \\
                        Slogans & 13 & 7 & 7 & 86 & 21 & 5 & 10 & 3 & 6 \\
                        Conversation Killer & 92 & 35 & 15 & 96 & 25 & 16 & 11 & 2 & 6 \\
                        \midrule
                        Loaded Language & 223 & 93 & 37 & 294 & 58 & 44 & 122 & 44 & 63 \\
                        Obfuscation-Vagueness-Confusion & 102 & 27 & 15 & 49 & 12 & 11 & 30 & 19 & 19 \\
                        Exaggeration-Minimisation & 87 & 48 & 21 & 75 & 22 & 21 & 37 & 15 & 20 \\
                        Repetition & 82 & 37 & 16 & 176 & 53 & 45 & 8 & 2 & 3 \\
                        \midrule
                        \midrule
                        Sentences & 3138 & 1346 & 500 & 2809 & 703 & 460 & 1101 & 473 & 500 \\
                        With any PT & 1548 & 676 & 267 & 1496 & 362 & 222 & 654 & 261 & 303 \\
                        Without any PT & 1590 & 670 & 233 & 1313 & 341 & 238 & 447 & 212 & 197 \\
                        \bottomrule
                \end{tabular}
        }
    \caption{Number of sentences with each type of PT across train-validation-test split.}
    \label{tab:dataset_stats_sentence_level}
        
\end{table}

\myparagraph{Results}
Table~\ref{tab:sentence-classification} presents results achieved by transformer-based classifiers\footnote{\href{https://huggingface.co/collections/SlavicNLP/persuasiontechniques-lrec2026}{huggingface.co/collections/SlavicNLP/\\persuasiontechniques-lrec2026}}.  The low performance for Russian may be attributable to the limited data relative to other languages---approximately 1,000 sentences, compared to 3,000 for Polish and Bulgarian in the train subset (see Table~\ref{tab:dataset_stats_sentence_level}).  Consequently, the setup for Russian was adjusted to train on the combined training and validation subsets. The final model, after 60 epochs, achieves modest recognition for some techniques: \emph{Name-Calling, Labeling} ($F_1$ = .27), \emph{Appeal to Values} ($F_1$ = .27), \emph{Causal Oversimplification} ($F_1$ = .26).

For Bulgarian and Polish, each with approximately 3,000 instances, the most frequent techniques yielded relatively high $F_1$ values: \emph{Questioning the Reputation} (.42 for Bulgarian; .40 for Polish), \emph{Loaded Language} (.42 for Bulgarian; .59 for Polish).  Certain techniques performed better in one language than in others---which is a pattern that correlates with the frequency of instances: for Bulgarian, \emph{Doubt} (.53) and \emph{Consequential Oversimplification} (.42); for Polish, \emph{Appeal to Values} (.61) and \emph{Appeal to Authority} (.61).

\begin{table}[t]
  \centering
     \resizebox{.99\columnwidth}{!}{%
    \begin{tabular}{@{} llrlrlr}
      \toprule
      & \multicolumn{2}{c}{\sc BG} & \multicolumn{2}{c}{\sc PL} 
      & \multicolumn{2}{c}{\sc RU} \\
      \midrule
Technique & $F_1$ & sup & $F_1$ & sup & $F_1$ & sup \\
      \midrule   
Questioning the Reputation            & {\sethlcolor{GreenYellow}\hl{.42}} & 41 & {\sethlcolor{GreenYellow}\hl{.40}} & 33 & .0 & 4 \\
Name-Calling, Labeling                & {\sethlcolor{LimeGreen}\hl{.68}} & 38 & .27 & 11 & .27 & 22 \\
Guilt by Association                  & .0 & 2 & .0 & 4 & .0 & 4 \\
Doubt                                 & {\sethlcolor{LimeGreen}\hl{.53}} & 62 & .22 & 26 & .15 & 20 \\
Appeal to Hypocrisy                   & {\sethlcolor{Yellow}\hl{.33}} & 26 & {\sethlcolor{Yellow}\hl{.35}} & 15 & .0 & 0\\
\midrule
Appeal to Popularity                  & .0 & 3 & .17 & 8 & .0 & 5 \\
Appeal to Values                      & .0 & 8 & {\sethlcolor{LimeGreen}\hl{.61}} & 19 & .27 & 18\\
Appeal to Authority                   & .0 & 6 & {\sethlcolor{LimeGreen}\hl{.61}} & 12 & .0 & 4 \\
Appeal to Fear/Prejudice              & {\sethlcolor{GreenYellow}\hl{.40}} & 13 & {\sethlcolor{GreenYellow}\hl{.41}} & 32 & .0 & 12 \\
Flag Waving                           & {\sethlcolor{GreenYellow}\hl{.40}} & 10 & .17 & 17 & .11 & 16 \\
\midrule					  
Causal Oversimplification             & .11 & 10 & .0 & 2 & .26 & 36 \\
False Dilemma-No Choice               & .20 & 12 & .0 & 3 & .0 &  17 \\
Consequential Oversimplification      & {\sethlcolor{GreenYellow}\hl{.42}} & 15 & .0 & 6 & .0 &  3 \\
False Equivalence                     & .0 & 8 & .0 & 4 & .0 & 3 \\
\midrule
Straw Man                             & .29 & 10 & .0 & 6 & .0 & 10 \\
Whataboutism                          & .17 & 8 & .0 & 3 & .0 &  5 \\
Red Herring                           & {\sethlcolor{GreenYellow}\hl{.40}} & 15 & {\sethlcolor{GreenYellow}\hl{.40}} & 7 & .0 & 10 \\
Appeal to Pity                        & .0 & 8 & {.29} & 8 & .0 & 13 \\
\midrule
Appeal to Time                        & .0 & 1 & {\sethlcolor{Yellow}\hl{.33}} & 8 & .0 & 3 \\
Slogans                               & .0 & 7 & .0 & 5 & .0 & 6 \\
Conversation Killer                   & {\sethlcolor{Yellow}\hl{.36}} & 15 & {\sethlcolor{Yellow}\hl{.36}} & 16 & .0 & 6 \\
\midrule
Loaded Language                       & {\sethlcolor{GreenYellow}\hl{.42}} & 37 & {\sethlcolor{LimeGreen}\hl{.59}} & 44 & .17 & 63 \\
Obfuscation-Vagueness-Confusion       & .08 & 15 & .0 & 11 & .0 & 19 \\
Exaggeration-Minimization             & .21 & 21 & .09 & 21 & .0 & 20 \\
Repetition                            & .17 & 16 & .21 & 45 & .0 & 3 \\
\midrule
Without any PT                        & .64 & 259 & .63 & 246 & .57 & 198 \\
      \bottomrule
ALL (micro)                           & .46 & 666 & .44 & 612 & .28 &  527 \\
ALL (macro)                           & .24 & 666 & .23 & 612 & .07 &  527 \\
     \bottomrule 
    \end{tabular}
    }
    \caption{Sentence level classification---$F_1$ and support for each class---for BERT-architecture models.  Scores above 0.5, in range 0.4--0.5, and in range 0.3--0.4  highlighted in green, lime green, and yellow, respectively.}
  \label{tab:sentence-classification}
\end{table}





\section{Conclusion and Future Work}
\label{sec:conclusions}


This paper describes a new corpus of excerpts from parliamentary debates in Bulgarian and Polish, and social media texts in Russian, annotated with fine-grained persuasion techniques at the text-span level and at the sentence level.  The corpus covers highly controversial topics discussed in both national and global discourse.  The paper gives an overview of the corpus creation process, detailed statistics about the datasets, and an analysis of the correlation of topic and persuasion techniques.  It presents baseline models for the task of persuasion technique detection and classification at the text-span and sentence level, and reports on the performance of these models in order to demonstrate the complexity of these tasks.  We make the corpus freely available for research purposes.
e
We believe that together with the accompanying baseline models, it will be useful for the community working on the detection of manipulative content in general, and in Slavic languages in particular.

In the future, we plan to: (a) expand the size of the corpus, (b) cover a wider range of Slavic languages, (c) improve the models by generating synthetic data, and (d) explore in depth the detection of the more challenging persuasion techniques, in particular, those that fall under the strategies of \emph{Simplifications} and \emph{Distractions}.

\section*{Acknowledgments}

Work carried out in Bulgaria was partially supported by the European Union---NextGenerationEU, through the National Recovery and Resilience Plan of the Republic of Bulgaria [Grant Project No. BG-RRP-2.004-0008].   

Work carried out in Finland was supported in part by the Research Council of Finland, Project {\em ``Know-AI''} (Grant 1359285), and by European Regional Development Fund (ERDF) Project {\em ``Generative AI and Knowledge Management''} (Grant 4740347).  We are grateful for their support.

\section*{Ethics}

\paragraph{Intended Use and Misuse Potential:}

The corpus presented in this paper was created to foster research on detection and classification of persuasion techniques for the domain of parliamentary debates 
and social media.  Given the potential risks of exploiting this corpus to produce manipulative content, we strongly advise responsible use of the data.

\paragraph{Fairness:}

The annotators were either: (a) researchers from the institutions of the authors of this manuscript, (b) students from the respective academic organizations, or (c) experts from a contracted professional annotation company. The annotators in the first two groups were fairly remunerated as part of their job, whereas the experts in the third group were compensated at rates based on their country of residence. 

\section*{Limitations}

\paragraph{Dataset Representativeness:}

The corpus covers parliamentary debates and propaganda narratives in various countries, and we strove to include utterances of speakers covering a wide political spectrum in each of these countries.  However, we must emphasize that these datasets should not be considered representative of the political landscape in any specific country or region, nor should they be considered as balanced in any way.

\paragraph{Biases:}

We have invested significant effort in training the annotators and acquainting them with the specifics of the persuasion technique taxonomy. Furthermore, cross-language quality control mechanisms have been implemented to ensure the highest quality of annotations. Nevertheless, some degree of intrinsic subjectivity might be present in the datasets. Therefore, models trained using these datasets might exhibit certain biases.

\section*{References}

\bibliographystyle{lrec2026-natbib}
\bibliography{custom}

\appendix

\section{Persuasion Technique Definitions}
\label{pt-definitions}


Below, we provide definitions of the persuasion techniques accompanied by examples in English (in blue) and in the Slavic languages (in brown) of the Shared Task. The text fragments highlighted in bold are the text spans to be annotated according to the guidelines presented in~\cite{JRC132862}.

The fine-grained techniques are grouped under 6 classes---the persuasion strategies.  The definitions of the persuasion techniques are taken directly from the Annex of~\cite{piskorski-etal-2023-multilingual}, with two new persuasion techniques, \textit{Appeal to Pity} and \textit{False Equivalence}, added for this task.

\subsection{Persuasion Strategy: Attack on Reputation}

\noindent \textbf{Name Calling or Labeling:}
a form of argument in which loaded labels are directed at an individual or a group, typically in an insulting or demeaning way. An object is labeled as something the target audience fears, hates, or, on the contrary, finds desirable or loves. This technique calls for a qualitative judgement that disregards facts and focuses solely on the essence of the subject being characterized. This technique is also in a way manipulative wording, as it appears as a nominal group rather than being a full-fledged argument with a premise and a conclusion. For example, in political discourse, typically one uses adjectives and nouns as labels that refer to political orientation, opinions, personal characteristics, and association to some organisations, as well as insults.  What distinguishes it from \emph{Loaded Language} (see~\ref{sec-pt-mw}), is that it is concerned only with the characterization of the subject.


\enExample{\annotated{'Fascist' Anti-Vax} Riot Sparks COVID Outbreak in Australia.}


\slavicExample{Trzeba zrozumieć, że bronią także i polskich granic przeciwko \annotated{rosyjskiemu   imperializmowi}, którego ducha wskrzesił Władimir Putin---\annotated{prezydent zbrodniarz}.} \translation{It is necessary to understand that they are also defending the Polish borders against \annotated{Russian imperialism}, whose spirit has been revived by Vladimir Putin---\annotated{the criminal president.}} 

\vspace{0.1cm}

\noindent \textbf{Guilt by Association:} Attacking an opponent or an activity by associating it with another group, activity, or concept that has sharply negative connotations for the target audience.  The most common example, which has given its name in the literature to this technique (i.e., \emph{Reduction ad Hitlerum}) is making comparisons with Hitler and the Nazi regime.  However, it is important to emphasize, that this technique is not restricted to comparisons to that group only. More precisely, this can be done by claiming a link or an equivalence between the target of the technique and any individual, group, or event in the present or in the past, which is or was negatively perceived (e.g., was considered a failure), or is depicted in such a way.

\enExample{\annotated{Manohar is a big supporter for equal pay for equal work. This is the same policy that all those extreme feminist groups support.} Extremists like Manohar should not be taken seriously.}

\cyrillicExample{Мы часто забываем, что после Второй мировой наши типа союзники, фран\-цузы \annotated{(на самом деле настоящие союзники Гитлера)}, стали срочно восстанавливать свою империю.} 
\translation{We often forget that after WWII our so-called allies, the French \annotated{(in fact, Hitler's allies)}, immediately started rebuilding their empire.}

\vspace{0.1cm}

\noindent \textbf{Casting Doubt:} Casting doubt on the character or the personal attributes of someone or something in order to question their general credibility or quality, rather than using a proper argument relevant to the topic.  This can be done for instance, by speaking about the target's professional background, as a way to discredit their argument.  Casting doubt can also be done by referring to some actions or events carried out or planned by some entity that are/were not successful, or appear as resulting in not achieving the planned goals.

\enExample{This task is quite complex. \annotated{Is his professional background, experience and the time left sufficient to accomplish the task at hand?}}

\cyrillicExample{\annotated{Има един-единствен корес\-пон\-дент от българска страна и по нашите медии се твърди, че те били обективни, представяли реално гледната точка, ка\-звали истината и прочее.}} \translation{\annotated{There was only one reporter from Bulgaria, and our media claimed that they were objective, presented a realistic point of view, told the truth, and so on.}}

\vspace{0.1cm}

\noindent \textbf{Appeal to Hypocrisy:} The reputation of the target is attacked by charging them with hypocrisy or inconsistency.  This can be done explicitly by calling out hypocrisy directly, or implicitly by underlining the contradictions between different positions that were held or actions that were done in the past.  A common way of calling out hypocrisy is by saying that someone who criticizes you for something you have done, has done it himself in the past.

\enExample{\annotated{How can you demand that I eat less meat to reduce my carbon footprint if you yourself drive a big SUV and fly for holidays to Bali?}}

\cyrillicExample{\annotated{Иначе СЕМ твърди, че е безпри\-страстен, но когато става въпрос за безпочвени обвинения към Русия или ма\-нипулиране на общественото мнение по този начин, някак си СЕМ пропуска това.}} \translation{\annotated{Otherwise, the CEM claims to be impartial, but when it comes to groundless accusations against Russia or manipulating public opinion in this way, the CEM somehow misses the mark.}}

\vspace{0.1cm}

\noindent \textbf{Questioning the Reputation:} This technique is used to attack the reputation of the target by making strong negative claims about it, focusing on undermining its character and moral stature rather than relying on an argument about the topic. Whether the claims are true is irrelevant for the effective use of this technique.  Smears can be used at any point in a discussion. One way of using this technique is to preemptively call into question the reputation/credibility of an opponent, before he has a chance to express himself, therefore biasing the audience's perception.  Hence, one of the names for this technique is ``poisoning the well.''

The main difference between \emph{Casting Doubt} (above) and \emph{Questioning the reputation} is that the former focuses on questioning the capacity, capabilities, and credibility of the target, while the latter aims to undermine the overall reputation, moral qualities, behaviour, etc.

\enExample{I hope I presented my argument clearly.  Now, \annotated{my opponent will attempt to refute my argument by his own fallacious, incoherent, illogical version of history.}}

\slavicExample{\annotated{Jedni i drudzy rządziliście Polską od 20 lat i nie przygotowaliście nas do obrony na czas wojny.}} \translation{\annotated{Together, you have governed Poland for 20 years and failed to prepare us for defense in times of war.}}

\subsection{Persuasion Strategy: Justification}

\noindent \textbf{Flag Waving:} Justifying or promoting an idea by appealing to the pride of a group or highlighting the benefits for that specific group. The stereotypical example would be national pride, and hence the name of the technique; however, the target may be any group, e.g., related to race, gender, political preference, etc. The connection to nationalism, patriotism, or benefit for an idea, group, or country might be inappropriate and is usually based on the presumption that the recipients already hold certain beliefs, biases, and prejudices about the given issue.  It can be seen as an appeal to emotions instead to logic of the audience aiming to manipulate them to win an argument. As such, this technique can also appear outside well-constructed arguments, by making statements that resonate with the particular group and as such setting up a context for further arguments.

\enExample{\annotated{We should make America great again, and restrict the immigration laws.}}

\slavicExample{\annotated{Wolna Ukraina i silna Unia Europejska, silna Polska stanowią podstawę polskiej racji stanu, to podstawa naszego bezpieczeństwa.}} \translation{\annotated{A free Ukraine and a strong European Union, a strong Poland, are the foundation of the Polish national interest, they are the basis of our security.}}

\vspace{0.1cm}

\noindent \textbf{Appeal to Authority:} attempting to add weight to an argument, an idea or information by simply stating that a particular entity considered to be an authority is the source of the information.  The entity mentioned as an authority may, but does not need to be, an actual authority in the specific domain to discuss a particular topic or to serve as an expert.  What is important, and makes it different from simply sourcing information, is that the tone of the text capitalizes on the weight of the alleged authority in order to justify some claim or conclusion. Referencing a valid authority is not a logical fallacy, while referencing an invalid authority is a logical fallacy, and both are captured within this label. In particular, a self-reference as an authority falls under this technique as well.

\enExample{\annotated{Since the Pope said that this aspect of the doctrine is true we should add it to the creed.}}

\cyrillicExample{\annotated{Глава ЦБ РФ Эльвира На\-би\-уллина назвала новые реалии тек\-то\-ническими изменениями в ми\-ро\-вой торговле, и с учётом всех нюансов происходящего это ещё очень деликатная формулировка.}} \translation{\annotated{The head of the Central Bank of Russia Elvira Nabiullina called the new situation a “tectonic shift in global trade,” and considering all the nuances of what is happening, this is still a very delicate formulation.}}

\vspace{0.1cm}

\noindent \textbf{Appeal to Popularity:} This technique gives weight to an argument or idea by justifying it on the basis that allegedly ``\textit{everyone}'' (or the vast majority) agrees with it, or ``\textit{nobody}'' disagrees with it.  The target audience is encouraged to gregariously adopt the same idea by considering ``\textit{everyone}'' as an authority, and to join in and take the same course of action. Here, ``\textit{everyone}'' might refer to the general public, key entities and actors in a certain domain, countries, etc.  Analogously, an attempt to persuade the audience not to do something because ``\textit{nobody else is taking the same action}'' falls under our definition of {\em Appeal to Popularity}.

\enExample{\annotated{Because everyone else goes away to college, it must be the right thing to do.}}

\cyrillicExample{\annotated{По предната точка Ви казах за последното социологическо проучване, в което 78\% от българските граждани не искат да се предоставя оръжие на Украйна, а Вие правите точно това, което не искат българските граждани.}} \translation{\annotated{In the previous point, I told you about the latest sociological survey, in which 78\% of Bulgarian citizens do not want weapons to be provided to Ukraine, and you are doing exactly what Bulgarian citizens do not want.}}

\vspace{0.1cm}

\noindent \textbf{Appeal to Values:} This technique gives weight to an idea by linking it to values seen by the target audience as positive. These values are presented as an authoritative reference in order to support or to reject an argument. Examples of such values are, for instance: tradition, religion, ethics, age, fairness, liberty, democracy, peace, transparency, etc.  When such values are mentioned outside the context of a proper argument by simply using certain adjectives or nouns as a way of characterizing something or someone, such references fall under another label, namely, \emph{Loaded Language}, which is a form of \emph{Manipulative Wording} (see~\ref{sec-pt-mw}).

\enExample{\annotated{It's standard practice to pay men more than women so we'll continue adhering to the same standards this company has always followed.}}

\cyrillicExample{\annotated{ В очередной раз удар нанесён по одной из самых чувствительных сфер---религиозным правам и свободам.}} \translation{\annotated{Another attack has been made on one of the most sensitive areas---religious rights and freedoms.}}

\vspace{0.1cm}

\noindent \textbf{Appeal to Fear, Prejudice:} This technique aims at promoting or rejecting an idea through the repulsion or fear the audience feels toward this idea (e.g., via exploiting some preconceived judgements) or toward its alternative. The alternative could be the status quo, in which case the current situation is described in a scary way with \emph{Loaded Language}. If the fear is linked to the consequences of a decision, it is often the case that this technique is used simultaneously with \emph{Appeal to Consequences} (see Simplification techniques in~\ref{sec-pt-s}), and if there are only two alternatives that are stated explicitly, then it is used simultaneously with the \emph{False Dilemma} technique (see~\ref{sec-pt-s}).

\enExample{It is a great disservice to the Church to maintain the pretense that there is nothing problematical about Amoris laetitia. \annotated{A moral catastrophe is self-evidently underway} and it is not possible honestly to deny its cause.}

\cyrillicExample{Много, много други такива неща са се случвали и за съжаление, \annotated{ние отиваме по едни стъпки, които са изключително опасни, изключително наистина трево\-ж\-ни за бъдещето на нашата държава.}} \translation{Many, many other such things have happened, and unfortunately,\annotated{we are taking extremely dangerous steps, extremely worrying for the future of our country.}}

\subsection{Persuasion Strategy: Distraction}

\noindent \textbf{Strawman:} This technique consists in creating an illusion of refuting the argument of the opponent's proposition, while the real subject of the argument was not addressed or refuted, but instead replaced with a false one. Often, this technique is referred to as a misrepresentation of the argument. First, a new argument is created via the covert replacement of the original argument with something that appears related, but is actually a different, distorted, exaggerated, or misrepresented version of the original proposition, which is referred to as ``\textit{setting up a strawman}.'' Subsequently, the newly created `\textit{false} argument (strawman) is refuted, which is referred to as ``\textit{knocking down the strawman}.'' Often, the strawman argument is created in such a way that it is easier to refute, and thus, creating the illusion of having defeated an opponent's real proposition. Fighting a strawman is easier than fighting a real person, which explains the name of this technique. In practice, it appears often as an abusive reformulation or explanation of what the opponent \textit{actually} means or intends.

\enExample{Referring to your claim that providing medicare for all citizens would be costly and a danger to the free market, I infer \annotated{that you don't care if people die from not having healthcare, so we are not going to support your endeavour}.}

\cyrillicExample{Има огромно значение, господин Иванов, дали българското знаме е отляво, или отдясно. Това нещо го знаете по протокол. \annotated{Ако казвате, че няма значение, това означава, че за Вас няма значение какъв точно ще бъде статутът на българското знаме в България, статутът на българския държавен герб и къде точно ще се полага}} \translation{It makes a huge difference, Mr Ivanov, whether the Bulgarian flag is on the left or the right. You know this from protocol.\annotated{If you say that it does not matter, it means that it does not matter to you exactly what the status of the Bulgarian flag will be in Bulgaria, the status of the Bulgarian state coat of arms and exactly where it will be placed.}}

\vspace{0.1cm}

\noindent \textbf{Red Herring:} This technique consists in diverting the attention of the audience from the main topic being discussed, by introducing another topic. The aim of attempting to redirect the argument to another issue is to focus on something the person doing the redirecting can better respond to or to leave the original topic unaddressed. The name of that technique comes from the idea that a fish with a strong smell (such as a herring) can be used to divert dogs from the scent of someone they are following.  A strawman (defined earlier) is a specific type of a red herring in that it distracts from the main issue by presenting the opponent's argument in an inaccurate light. 

\enExample{Lately, there has been a lot of criticism regarding the quality of our product. \annotated{We've decided to have a new sale in response, so you can buy more at a lower cost!}.}

\cyrillicExample{Недавно она прочитала лекцию о необходимости войны с ухоженным газоном, потому что «это символ сексизма, расизма и экологического разрушения». \annotated{Среди друзей Аджубей много проукраинских активистов и адептов движений Black lives matter и ЛГБТ.}} \translation{She recently gave a lecture on the need for a war on manicured lawns because “they are a symbol of sexism, racism and ecological destruction” \annotated{Adzhubey's friends include many pro-Ukrainian activists and adherents of the Black lives matter and LGBT movements.}}

\vspace{0.1cm}

\noindent \textbf{Whataboutism:} Attempt to discredit an opponent's position by charging them with hypocrisy without directly disproving their argument.  Rather than answering a critical question or argument, an attempt is made to retort with a critical counter-question that expresses a counter-accusation, e.g., mentioning double standards, etc.  The intent is to distract from the content of a topic and to actually switch the topic. There is a fine distinction between this technique and \emph{Appeal to Hypocrisy}, introduced earlier: the former is an attack on the argument and introduces irrelevant information to the main topic, while the latter is an attack on reputation and highlights the hypocrisy of double standards on the same or a closely related topic.

\enExample{\annotated{A nation deflects criticism of its recent human rights violations by pointing to the history of slavery in the United States.}}

\cyrillicExample{Добре, на Хърватия е пораснал---окей. \annotated{А Естония и Финландия, които са на минус, и Ирландия, които са в евро\-зоната, какво правим?}} \translation{Okay, Croatia's has grown---okay. \annotated{And what about Estonia and Finland, which are in the red, and Ireland, which are in the eurozone, what do we do?}}

\vspace{0.1cm}

\noindent \textbf{Appeal to Pity:} Evokes feelings of pity, sympathy, compassion or guilt in audience to distract it from focusing on evidence, rational analysis and logical reasoning, so that it accepts the speaker's conclusion as truthful solely based on these emotions.  It is an attempt to sway opinions and fully substitute logical evidence in an argument with a claim intended to elicit pity or guilt.

\enExample{\annotated{If this person is found guilty of this crime, his ten children will be left without a parent at home, therefore the jury must submit a verdict of innocence.}}

\cyrillicExample{\annotated{Напуганные, изнурённые отсутствием спокойствия и элемен\-тарных условий для жизни}, женщины всё равно не были сломлены и не потеряли надежду на освобождение российскими подразделениями их родного хутора.} \translation{\annotated{Frightened, exhausted by the insecurity and lack of basic living conditions}, the women were still not broken and did not lose hope for the liberation of their village by Russian troops}

\subsection{Persuasion Strategy: Simplification}
\label{sec-pt-s}

\noindent \textbf{Causal Oversimplification:} Assuming a single cause or reason when there are actually multiple causes for an issue.
This technique has the following logical form(s): (a) \textit{Y occurred after X; therefore, X was the only cause of Y}, or (b) \textit{X caused Y; therefore, X was the only cause of Y (although A, B, C...etc. also contributed to Y.)}

\enExample{School violence has gone up and academic performance has gone down since video games featuring violence were introduced. \annotated{Therefore, video games with violence should be banned, resulting in school improvement.}}

\cyrillicExample{И если собственно украинские возможности к сопротивлению закончились к концу марта 22го, что и привело Киев к Стамбулу, \annotated{то выигрыш 3х недель, вселил уверенность в запад, и поэтому было решение Джонсона, продолжать войну.}} \translation{If Ukraine's own capacity for resistance had run out by the end of March 22, bringing Kyiv to Istanbul, then \annotated{the three-week gain gave the West confidence, and that is why Johnson decided to continue the war.}}

\vspace{0.1cm}

\noindent \textbf{False Dilemma or No Choice:} Sometimes called the \textit{either-or} fallacy, a false dilemma is a logical fallacy that presents only two options or sides when there are actually many. One of the alternatives is depicted as a \textit{no-go} option, hence the only choice is the other option. In extreme cases, the author tells the audience exactly what actions to take, eliminating any other possible choices (also referred to as \textit{Dictatorship}).

\enExample{\annotated{There is no alternative to Pfizer Covid-19 vaccine. Either one takes it or one dies.}}

\slavicExample{\annotated{Debatujemy dzisiaj o bardzo ważnej i – ośmielę się powiedzieć – kluczowej dla nas sprawie, sprawie życia i śmierci.}} \translation{\annotated{Today we are debating a very important and, I dare say, crucial issue for us, a matter of life and death.}}

\vspace{0.1cm}

\noindent \textbf{Consequential Oversimplification:} An argument or an idea is rejected and instead of discussing whether it makes sense and/or is valid, the argument affirms, without proof, that accepting the proposition would imply accepting other propositions that are considered negative. This technique has the following logical form: \textit{if A will happen then B, C, D, ... will happen}. The core essence behind this fallacy is an assertion one is making of some `\textit{first}' event/action leading to a domino-like chain of events that have some significant negative effects and consequences that appear to be ludicrous. This technique is characterized by {\em ignoring and/or understating the likelihood of the sequence of events from the first event leading to the end point} (last event). In order to take into account symmetric cases, i.e., using \emph{Consequential Oversimplification} to promote or to support certain action in a similar way, we also consider cases when the sequence of events leads to positive outcomes (i.e., 
encouraging people to undertake a certain course of action(s), with the promise of a major positive event in the end).

\enExample{\annotated{If we begin to restrict freedom of speech, this will encourage the government to infringe upon other fundamental rights, and eventually this will result in a totalitarian state where citizens have little to no control of their lives and decisions they make.}}

\cyrillicExample{\annotated{Сокрытие правды и подмена понятий приведет к тому, что управлять умами и историей будет противник на нашей территории, выдавая правду с нужным ему уклоном.}} \translation{\annotated{Concealing the truth and substituting concepts will result in the enemy controlling minds and history on our territory, spreading the truth with an intended bias.}}

\vspace{0.1cm}

\noindent \textbf{False Equivalence:} A technique that attempts to treat scenarios that are significantly different as if they had equal merit or significance. In particular, an emphasis is placed on one specific shared characteristic between the items of comparison in the argument that is off by an order of magnitude, oversimplified, or important additional factors have been ignored.  The introduction of certain shared characteristics of the scenarios is then used to consider them equivalent. This technique has the following logical form: \textit{A and B share some characteristic X. Therefore, A and B are equivalent}.

\enExample{\annotated{The introduction or restrictive hours of alcohol sales boosted the black market industry, and analogously, one can expect that the introduction of too restrictive anti-abortion regulations will lead to growth of the illegal abortion business}.}

\slavicExample{\annotated{To właśnie Führer jako pierwszy wprowadził wolną aborcję dla Polek oraz dla innych kobiet z narodów podbitych. Chodziło o fizyczne zniszczenie ludności niearyjskiej i zdobycie lebensraumu dla Niemców. Hitler rozumiał, że jeśli zalegalizuje aborcję, stanie się ona zjawiskiem masowym i spowoduje spadek urodzeń. Na ziemiach podbitych przez Niemcy dzieci niearyjskie uważano za zagrożenie, więc wdrażano politykę sprzyjającą aborcji. Równocześnie za to samo, za zabicie dziecka niemieckiego w Niemczech groziła kara śmierci. A dyktator groził: osobiście zastrzelę tego idiotę, który chciałby wprowadzić w życie przepisy zabraniające aborcji na wschodnich terenach okupowanych. Jaka jest analogia? Kto powiedział: każda odmowa aborcji będzie zgłaszana do prokuratury? Premier rządu rewolucji}} \translation{\annotated{It was the Führer who first introduced free abortion for Polish women and other women from conquered nations. The idea was to physically destroy the non-Aryan population and gain Lebensraum for the Germans. Hitler understood that if he legalized abortion, it would become a mass phenomenon and cause a decrease in births. In the lands conquered by Germany, non-Aryan children were considered a threat, so a policy favoring abortion was implemented. At the same time, the same thing, killing a German child in Germany, was punishable by death. And the dictator threatened: I will personally shoot this idiot who would want to implement regulations prohibiting abortion in the occupied eastern territories. What is the analogy? Who said: every refusal to have an abortion will be reported to the prosecutor's office? The prime minister of the government of the revolution}}

\cyrillicExample{В 1990-х годах были скинхеды---группы асоциальной молодежи, которые тол\-пой нападали на лиц неевропейской на\-руж\-ности, на так сказать «черных». \annotated{Теперь скин\-хеды---это группы асоциаль\-ной моло\-дежи среднеазиатской наруж\-ности, ко\-то\-рые так и не смогли гар\-монично жить рядом с русскими, и толпой из\-бивают русских парней и на\-силуют русских девочек}} \translation{In the 1990s, there were the skinheads---groups of antisocial youth who mobbed people of non-European appearance, the so-called “blacks”. \annotated{Now skinheads are groups of antisocial youth of Central Asian appearance, who failed to live peacefully next to Russians, and crowds of them beat up Russian guys and rape Russian girls}}

\subsection{Persuasion Strategy: Call}

\noindent \textbf{Slogans:} A brief and striking phrase that may include labeling and stereotyping. Slogans tend to act as emotional appeals.

\enExample{\annotated{Immigrants welcome, racist not!}}

\cyrillicExample{\annotated{Да живее България!}} \translation{\annotated{Long live Bulgaria!}}

\vspace{0.1cm}
 
\noindent \textbf{Conversation Killer:} This includes words or phrases that discourage critical thought and meaningful discussion about a given topic. They are a form of \emph{Loaded Language}, often passing as folk wisdom, intended to end an argument and quell cognitive dissonance.

\enExample{I'm not so na\"{i}ve or simplistic to believe we can eliminate wars. \annotated{You can't change human nature.}}

\slavicExample{\annotated{Takie są fakty i taka jest polska racja stanu.}} \translation{\annotated{These are the facts, and this is the Polish national interest.}}

\vspace{0.1cm}

\noindent \textbf{Appeal to Time:} The argument is centered around the idea that the time has come for a particular action. The very timeliness of the idea is part of the argument.

\enExample{This is no time to engage in the luxury of cooling off or to take the tranquilizing drug of gradualism. \annotated{Now is the time to make real the promises of democracy. Now is the time to rise from the dark and desolate valley of segregation to the sunlit path of racial justice.}}

\slavicExample{\annotated{Można powiedzieć, że w wielu wymiarach nastał czas prawdy.}} \translation{\annotated{It can be said that in many ways, the moment of truth has arrived.}}

\subsection{Persuasion Strategy: Manipulative Wording}
\label{sec-pt-mw}

\noindent \textbf{Loaded Language:} use of specific words and phrases with strong emotional implications (either positive or negative) to influence and to convince the audience that an argument is valid. It is also known as \emph{Appeal to Argument from Emotive Language}.

\enExample{They keep feeding these people with \annotated{trash}. They should stop.}

\slavicExample{\annotated{Nękanie} zasłużonej dla szerzenia polskości instytucji \annotated{bezzasadnymi} pozwami   odbierane jest m.in. przez moich wyborców jako działania mające na celu  \annotated{sparaliżowanie} funkcjonowania tej fundacji.} \translation{The \annotated{harassment} of an institution that has earned merit in promoting Polish identity through \annotated{groundless} lawsuits is perceived, among others by my constituents, as actions aimed at \annotated{paralyzing} the functioning of this foundation.}

\vspace{0.1cm}

\noindent \textbf{Obfuscation, Intentional Vagueness, Confusion:} This fallacy uses words that are deliberately unclear, so that the audience may have its own interpretations. For example, an unclear phrase with multiple or unclear definitions is used within the argument and, therefore, does not support the conclusion. Statements that are imprecise and intentionally do not fully or vaguely answer the posed question fall under this category.

\enExample{\annotated{Feathers cannot be dark, because all feathers are light!}}
\cyrillicExample{\annotated{Затем следует «вишенка»: дорогой доллар превращается в «пылесос» для капиталов со всего мира.
}} \translation{\annotated{Then comes the “cherry on top”: the expensive dollar turns into a “vacuum cleaner” for capital from all over the world.}}

\vspace{0.1cm}

\noindent \textbf{Exaggeration or Minimisation:} This technique consists of either representing something in an excessive manner---by making things larger, better, worse (e.g., \textit{the best of the best}, \textit{quality guaranteed})---or by making something seem less important or smaller than it really is (e.g., saying that an insult was just a joke), downplaying the statements and ignoring the arguments and the accusations made by an opponent.

\enExample{From the seminaries, to the clergy, to the bishops, to the cardinals, \annotated{homosexuals are present at all levels, by the thousand}.}

\slavicExample{\annotated{Europa prowadzi również najbardziej dramatyczną wojnę, wojnę demograficzną, którą przegrywa.}} \translation{\annotated{Europe is also fighting its most dramatic war, the demographic war, which it is losing.}}

\vspace{0.1cm}

\noindent \textbf{Repetition:} The speaker uses the same word, phrase, story, or imagery repeatedly in the hope that the repetition will persuade the audience.

\enExample{\annotated{Hurtlocker deserves an Oscar}. Other films have potential, but they do not \annotated{deserve an Oscar like Hurtlocker does}. The other movies may deserve an honorable mention but \annotated{Hurtlocker deserves the Oscar}.}

\cyrillicExample{И няма да бъдем готови, защото имаме структурни, чисто фундаментални проблеми и неща, които трябва да решим, \annotated{и това няма да се случи за 6 месеца, няма да се случи за година, няма да се случи и за две години.}} \translation{And we will not be ready, because we have structural, purely fundamental problems and issues that we need to resolve, \annotated{and this will not happen in six months, it will not happen in a year, it will not happen in two years.}}


\end{document}